%% file: arxiv-version.tex
\documentclass[a4paper, 10pt, conference]{ieeeconf}      
\usepackage{FG2024}
\usepackage{arydshln}
\usepackage{amsmath, amsfonts, times, epsfig, amssymb,  blindtext, float, subcaption, multicol, array, multirow, graphicx, etoolbox,  accents, tabularx, tabu, comment, soul}
\usepackage{color}
\usepackage{ifthen}
\usepackage{soul}
\usepackage{booktabs}
\usepackage{colortbl}
\def\etal{\emph{et al}.}
\def\ie{\emph{i.e.,}}
\def\eg{\emph{e.g.,}}
\def\sota{\emph{state-of-the-art}}
\definecolor{tablegray}{gray}{.9}
\makeatletter
\let\NAT@parse\undefined
\makeatother
\usepackage[backref=page,colorlinks=true]{hyperref}
\usepackage{cite}

\newboolean{highlightText}
\setboolean{highlightText}{false}
\newcommand{\myhighlight}[1]{\ifthenelse{\boolean{highlightText}}{\textcolor{red}{#1}}{#1}}
\newcommand{\mymovedparagraphs}[1]{\ifthenelse{\boolean{highlightText}}{\textcolor{blue}{#1}}{#1}}

\newcommand\blfootnote[1]{%
  \begingroup
  \renewcommand\thefootnote{}\footnote{#1}%
  \addtocounter{footnote}{-1}%
  \endgroup
}

\FGfinalcopy 
\IEEEoverridecommandlockouts                              
\def\FGPaperID{5} 

\title{\LARGE \bf Two Hands Are Better Than One: Resolving Hand to Hand Intersections via Occupancy Networks}

\author{\parbox{16cm}{\centering
    {\large Maksym Ivashechkin, Oscar Mendez, Richard Bowden}\\
    {\normalsize
    CVSSP, University of Surrey, Guildford, United Kingdom}\\
    {\normalsize \texttt {\{m.ivashechkin, o.mendez, r.bowden\}@surrey.ac.uk}}}%
}

\usepackage{fancyhdr}
\begin{document}

\ifFGfinal
\thispagestyle{empty}
\pagestyle{empty}
\else
\author{Anonymous FG2024 submission\\ Paper ID \FGPaperID \\}
\pagestyle{plain}
\fi
\maketitle
\thispagestyle{plain}
\pagestyle{plain}


\begin{abstract}
3D hand pose estimation from images has seen considerable interest from the literature, with new methods improving overall 3D accuracy.
One current challenge is to address \myhighlight{hand-to-hand interaction} where self-occlusions and finger articulation pose a significant problem \myhighlight{to estimation}.
Little work has \myhighlight{applied} physical constraints that minimize the hand intersections that occur as a result of noisy estimation.
This work \myhighlight{addresses} the intersection of hands by exploiting an occupancy network that represents the hand's volume as a continuous manifold.
This allows us to model the probability distribution of points being inside a hand.
We designed an intersection loss function to minimize the likelihood of hand-to-point intersections.
Moreover, we propose a new hand mesh parameterization that is superior to the commonly used MANO model \myhighlight{in many respects including} lower mesh complexity, underlying 3D skeleton extraction, watertightness, etc.
On the benchmark~\textsc{InterHand2.6M} dataset, the models trained \myhighlight{using our} intersection loss achieve better results than the~{\sota} by significantly decreasing the number of hand intersections while lowering the mean per-joint positional error.
\myhighlight{Additionally, we demonstrate superior performance for 3D hand uplift on \textsc{Re:InterHand} and \textsc{SMILE} datasets and show reduced hand-to-hand intersections for complex domains such as sign-language pose estimation.
}
\end{abstract}

\input{sections/introduction}

\input{sections/methodology}

\input{sections/experiments}
\input{sections/conclusions}

\blfootnote{\hspace{-0.3cm}{\bf Acknowledgement}. This work was supported by the SNSF project `SMILE II' (CRSII5 193686), European Union's Horizon2020 programme (`EASIER' grant agreement 101016982) and the Innosuisse IICT Flagship (PFFS-21-47). This work reflects only the authors view and the Commission is not responsible for any use that may be made of the information it contains.}

\clearpage
{\small
\bibliographystyle{ieee}
\bibliography{egbib}
}

\end{document}

%% file: sections/introduction.tex
\section{INTRODUCTION}
\myhighlight{Hand pose estimation in 3D has seen lots of interest across} a broad range of applications.
For instance, 3D hand estimation can be used for sign language recognition, human-computer interaction, animation, virtual/augmented reality, etc.
3D hand estimation is an extremely challenging problem due to factors including motion blur, self-occlusion of the fingers, and interaction with body and face.
The literature contains many works on single 3D hand pose estimation ({\eg} \cite{spurr2018cvpr, zb2017hand, spurr2021self, zhao_recover_3d, Chen2018GeneratingRT, disentagling_hands_linlin}) that tackle this problem by either direct hand prediction from an image or by decomposing estimation into an image-to-2D, then 2D-to-3D uplift.

The interaction of two hands in 3D space presents even greater complexity due to mutual occlusions and intersections.
A naive approach \myhighlight{is to apply a single-hand model twice to an image but this leads} to poor estimation that lacks realism, especially in subtle cases such as interlocking fingers.
\myhighlight{Furthermore}, interacting hands may provide additional information on mutual hand position, and this can prove beneficial in decreasing the search space of solutions.

\subsection{Related Work}
Early works on 3D pose estimation for interacting hands tried to solve the problem using classic optimization approaches.
Ballan~{\etal}~\cite{ballan_motion} proposed tackling the problem by exploiting salient points and formulating a differentiable objective function that incorporates edges, optical flow, and collisions extracted from an image.
Oikonomidis~{\etal}~\cite{Oikonomidis_tracking} tracked interacting hands by using a stochastic optimization method with the objective of finding the two-hand configuration that best explains observations from an RGB-D sensor.

With the rise of deep learning, hand interaction was approached by Taylor~{\etal}~\cite{Taylor_Articulated} who parameterized hands and exploited an articulated signed distance function to fit their model to multiview depth data.
Mueller~{\etal}~\cite{Mueller_realtime} proposed a method that uses a single depth camera along with an angular hand parameterization.
A multiview setup for interacting hand estimation from RGB images was employed by Smith~{\etal}~\cite{Smith_Constraining} where a physically based deformable model constrains a vision-based tracking algorithm to tackle self-occlusions and self-intersections.

\begin{figure*}[t]
    \centering
    \includegraphics[width=0.95\linewidth]{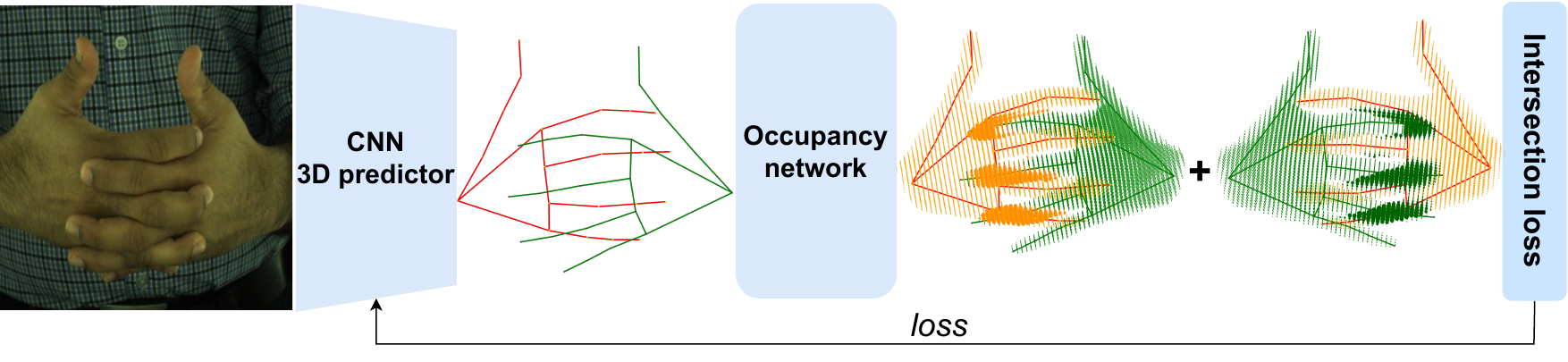}
    \caption{The figure demonstrates the pipeline for accurate 3D interacting hands estimation. The input image is processed via a CNN model ({\eg} ResNet~\cite{resnet}, MediaPipe, etc.) that enables the extraction of image features or 2D keypoints necessary to uplift hands into 3D. Note our approach is invariant to the backbone used. Afterward, a pre-trained occupancy network with frozen weights is conditioned via the right hand, and the intersections are tested with the left hand, and optionally \emph{vice-versa} with the hands flipped as illustrated. 
    The red and green edges highlight the right and left hands, respectively. Light green and light orange points visualize the density of hands determined by the occupancy network and their size emphasizes the likelihood of intersection. Since both hands are fully differentiable with respect to the occupancy and CNN networks, it provides efficient backpropagation of the intersection loss. The source image is taken from the~\textsc{InterHand2.6M} dataset.}
    \label{fig:pipeline}
\end{figure*}

More recent methods apply hand estimation to a single RGB image exploiting various techniques including image segmentation, mesh rendering, relative depth regression, etc.
\myhighlight{A common approach is to parameterize the hand using the MANO~\cite{MANO:SIGGRAPHASIA:2017} model, which represents the hand skeleton via joint orientations, and the} hand's volume is parameterized by a shape vector.
Moreover, the MANO library provides an efficient model that converts angular and shape hand parameterization into a 3D mesh surface.
MANO delivers many benefits that help to enforce a realistic hand, work with volumetric hand shapes, and render a mesh onto an image. 
But as will be seen, it also has shortcomings.

RGB2Hands~\cite{RGB2Hands} by Wang~{\etal} tackles 3D pose estimation and tracking of interacting hands from a monocular input by leveraging a segmentation mask, 2D detection, and dense matching to regress MANO hand parameters.
Zhang~{\etal}~\cite{Zhang2021InteractingT3} address the hand interaction problem by applying a hand pose-aware attention module to retrieve features corresponding to each hand.

The decomposition of interacting hands was tackled by Meng~{\etal}~\cite{meng2022hdr} who utilize de-occlusion and removal modules to recover the appearance content of the occluded part of one hand and remove the distracting hand.
On the contrary, Fan~{\etal}~\cite{fan2021digit} process an image using a per-pixel semantic part segmentation mask and a visual feature volume to leverage the per-pixel probabilities directly during pose estimation without decoupling the segmentation stage or individual hands in the pipeline.
Rong~{\etal}~\cite{rong2021ihmr} introduce a two-stage approach where at the first stage the CNN module makes a coarse prediction of interacting hands, and the second step progressively ameliorates the hands' collisions through a series of factorized refinements.

\textsc{InterHand2.6M}~\cite{Moon_2020_ECCV_InterHand2.6M} is a benchmark dataset for single and interacting hand pose estimation released by Moon~{\etal}. It was released together with a baseline approach that exploits a ResNet image model to regress 2.5D keypoints (image points with relative depth) that, via back-projection, are uplifted to 3D space.
This dataset has become widely used for hand pose evaluation and comparison.
The most recent~{\sota} methods achieve excellent performance, setting a high bar for interacting hand estimation on the \textsc{InterHand2.6M} dataset.
One of those methods is IntagHand~\cite{Li2022intaghand} by Li~{\etal} which uses a graph convolutional neural network to solve occlusions and interactions. \myhighlight{This is achieved} by adding attention-based modules to implicitly obtain vertex-to-image alignment that encode the coherence of interacting hands.
Jiang~{\etal} present the A2J-Transformer~\cite{jiang2023a2jtransformer} that extends the A2J framework~\cite{Xiong_a2j}, which uses anchor points to capture global-local spatial context.
The A2J-Transformer includes several advantages such as leveraging self-attention across local anchor points for spatial context awareness and utilizing anchor points as learnable queries with adaptive feature learning in 3D.
Yu~{\etal}~\cite{yu2023acr} introduce the ACR framework that explicitly mitigates interdependencies between hands by leveraging center and part-based attention for feature extraction and learns the corresponding cross-hand prior.

\mymovedparagraphs{A similar work that tackles hand-to-object interactions is HALO~\cite{karunratanakul2021halo} by Karunratanakul~{\etal} who present an occupancy network for continuous hand representation.
HALO employs a 3D sparse point cloud as input to the network to predict hand surface and utilizes hand-to-object occupancies to minimize the number of intersections.
However, it did not explore resolving hand-to-hand intersections.}

\subsection{Motivation \& Contributions}
\myhighlight{
Our primary focus is to improve 3D hand pose estimation from a single image by leveraging the physical constraints of hand-to-hand interaction.
Importantly, our framework can be applied to any \sota~3D hand estimation approach to improve performance. We demonstrate a reduction in hand intersections for 5~\sota~approaches. Through extensive evaluation on~\textsc{InterHand2.6M}~\cite{Moon_2020_ECCV_InterHand2.6M}, \textsc{Re:InterHand}~\cite{moon2023reinterhand}, and SMILE~\cite{ebling-etal-2018-smile} sign-language datasets, we show significant reduction in the number of intersections and 3D per-joint error.}

\myhighlight{
To do this we employ an occupancy network to exclude mutual intersections.
The occupancy model provides a continuous volumetric representation of the hand conditioned on a sparse skeletal model. We propose a new hand mesh parameterization that exploits a kinematic hand model which is more robust than the MANO framework. But this is only used to train the occupancy network. }

\myhighlight{
By leveraging the hand occupancy, we can resolve volumetric intersections without the use of a mesh.}
\myhighlight{We structure our approach around an end-to-end differentiable pipeline and employ an intersection loss that, in combination with a CNN, allows the reduction of hand intersections which improves 3D estimation.
In contrast to HALO, which minimizes hand-to-object intersections, we model intersections for two dynamic and interacting hands. This is a more significant challenge than hand-to-object, as it involves the simultaneous prediction of both articulated hands, while in hand-to-object, the object usually remains constant.}

%% file: sections/methodology.tex
\vspace{-0.2cm}
\section{METHODOLOGY}
\vspace{-0.35cm}
We propose an intersection loss for 3D interacting hand estimation that enforces physical constraints and visual realism.
The pipeline of our approach is outlined in Fig.~\ref{fig:pipeline}.
\myhighlight{The method relies on a CNN-based 3D pose estimator to lift initial 2D hand skeletons to 3D.}
By exploiting a pre-trained occupancy network~\cite{OccupancyNetworks} conditioned with a 3D skeleton, we apply an unsupervised intersection loss on both predicted hands.
The occupancy network is pre-trained on hand meshes extracted with a custom hand mesh parameterization.

\subsection{Hand Mesh Representation}
The MANO library is extensively used in the literature as a hand mesh representation.
It provides essential attributes such as differentiability, angular and shape mesh parameterization, facilitates retrieval of a 3D skeleton via linear blend skinning, and a heightened level of visual realism.
Unfortunately, despite all the advantages, it has certain limitations.
Firstly, it has a reliance on a pre-trained statistical hand model computed with principal component analysis.
Secondly, the 3D skeleton can only be obtained after mesh regression.
Finally, the generated mesh is not watertight.

The primary challenge lies in the complexity of fitting MANO meshes to pre-existing 3D skeletons,~{\ie} when 3D hand skeletons are known, but the goal is to obtain the volumetric shape.
This optimization could be done over the shape and angular MANO parameters.
However, the pre-computed MANO weights may not match the target skeleton distribution, which can result in strange mesh deformations.

Therefore, we present a new hand-mesh parameterization that is more practical for use with hand-to-hand interaction.
The core of the proposed mesh model is a 3D hand skeleton generated with forward kinematics combining joint angles and bone lengths.
We add vertices along the fingers that span the envelope of the hand, providing a volumetric shape.
Through an awareness of the arrangement of nodes, we are able to triangulate the hand's surface and construct a watertight mesh ({\ie} no holes) using triangular faces.
The full process of adding new nodes is fully automatic, and more vertices can be added to give a more complex hand shape.
The proposed custom mesh has multiple advantages over the MANO mesh:
\begin{enumerate}
    \item It is watertight.
    \item It has less than half the vertices (307 vs. 778).
    \item The volumetric shape is added on top of the skeleton (as opposed to MANO that generates the mesh and then skeleton), which enables lazy evaluation if only a skeleton is needed.
    \item No dependency on the pre-trained weights.
\end{enumerate}
The mesh watertightness plays a crucial role in training the occupancy network since it allows us to determine whether a 3D point is inside the mesh (using ray casting~\cite{ray_casting}).
Both MANO and the proposed meshes are differentiable, and the proposed model has more parameters to differentiate, which gives more user control over subtle details.

\begin{figure}[t]
    \centering
    \includegraphics[width=0.75\linewidth]{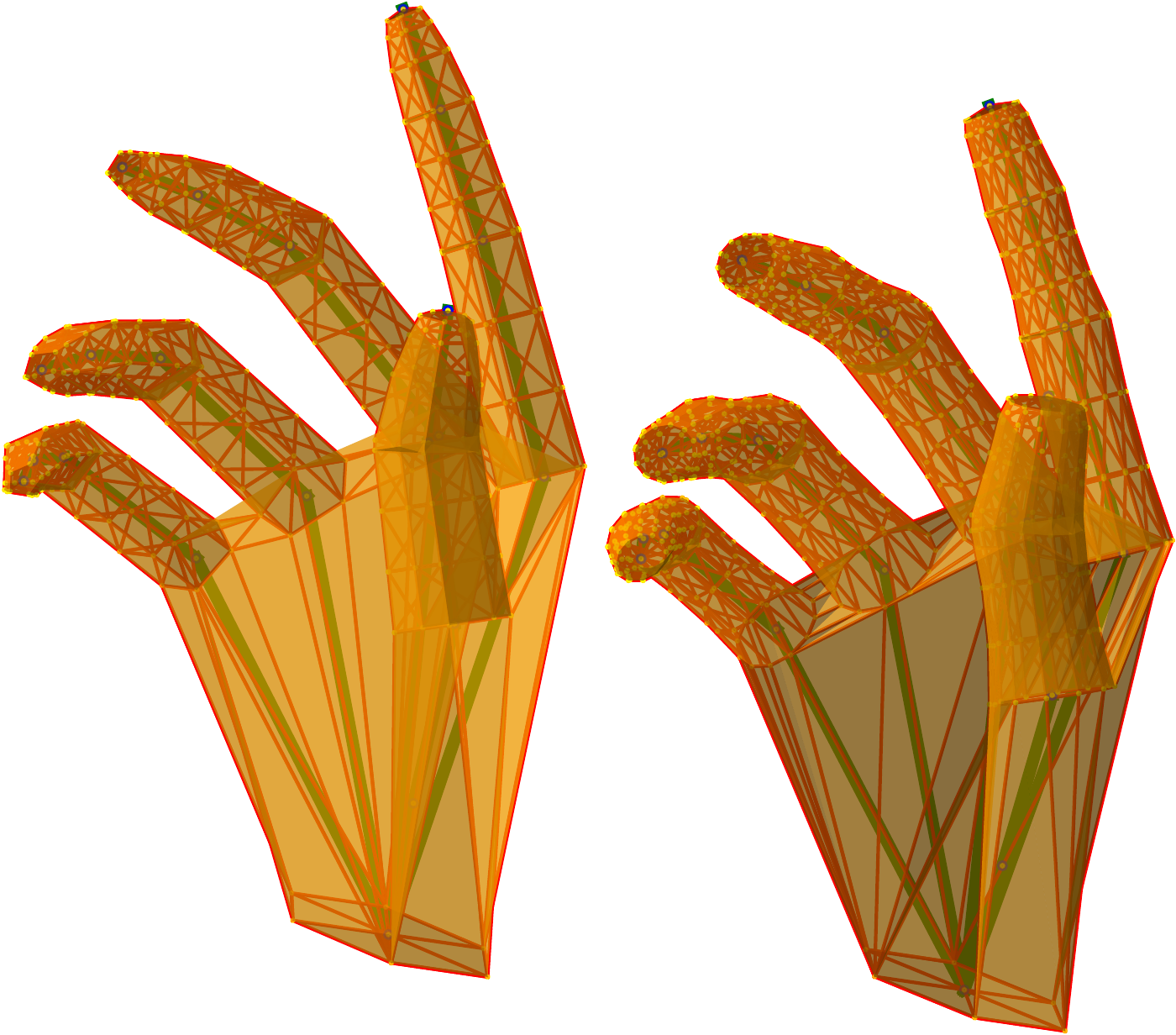}
    \vspace{-0.15cm}
    \caption{Comparison of plain (left) and complex (right) watertight hand meshes generated with our parameterized mesh model. The green pose (underneath the orange envelope) is found using forward kinematics (FK) that combine angles and bone length. The yellow points are also obtained via FK with pre-determined offsets from the underlying skeleton, the red triangles span the entire hand surface. }
    \label{fig:ik_mesh}
\end{figure}

To provide a good trade-off between visual quality and complexity, we designed a refined version of the proposed mesh model that has more points (699) and triangles, for the purpose of visualization.  
Fig.~\ref{fig:ik_mesh} demonstrates a visual comparison of two meshes.
MANO versus the proposed custom mesh comparison is illustrated in Fig.~\ref{fig:mesh-comparison}.

\subsection{Occupancy Network}
The occupancy network serves as the primary component in the proposed pipeline.
\myhighlight{Its purpose is to model} the volumetric shape of the hands and resolve intersections at the physical level, since it determines whether a point occupies the same space as a hand.
The benefit of utilizing an occupancy network is to obtain a continuous manifold of the hand that can be used in a lazy evaluation.
To obtain such a representation, the occupancy network is conditioned with a feature vector that corresponds to the desired shape.

Mathematically, let $\mathcal{O}: \mathbb{R}^3 \times \mathbb{F} \rightarrow [0, 1]$ be an occupancy network that given a point in 3D space ${\bf x} \in \mathbb{R}^3$, and feature vector ${\bf f} \in \mathbb{F}$, returns a probability $p \in [0,1]$ of the point ${\bf x}$ being occupied in the target space conditioned by vector ${\bf f}$.

Implementation-wise, the occupancy network consists of an encoder and a decoder.
The encoder processes 3D observations to produce a feature vector and acquires knowledge about the mean and standard deviation of the Gaussian distribution within the latent space.
At inference, the model samples the learned latent space, and the decoder transforms the latent space into occupancy logits, which can be converted to probabilities using the Bernoulli two-class model.

\vspace{-0.4cm} 
\subsection{Intersection Loss}
\begin{figure}
    \centering
    \begin{subfigure}{0.32\linewidth}
    \includegraphics[width=0.99\linewidth]{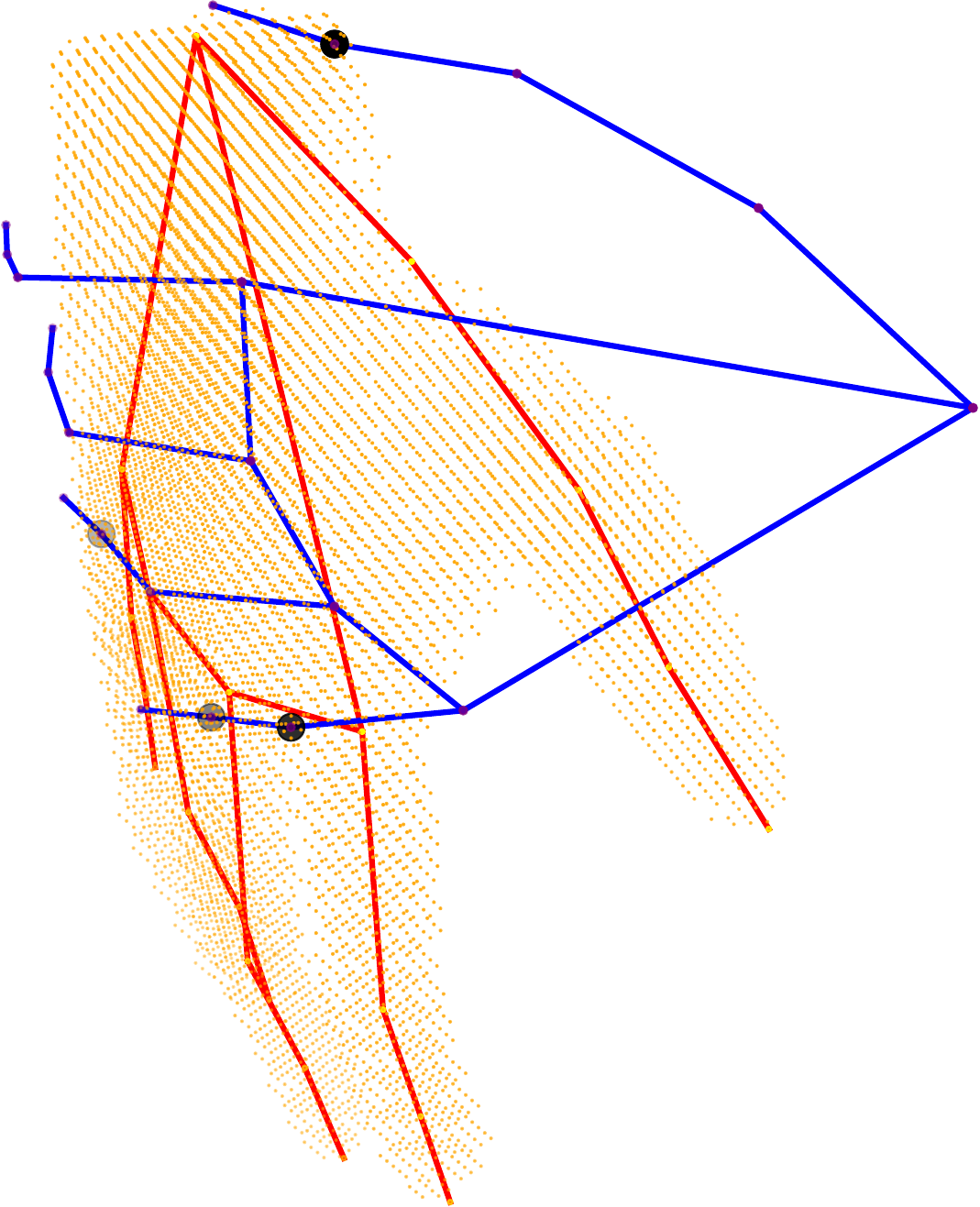}
    \caption{sparse}
    \end{subfigure}
    \begin{subfigure}{0.32\linewidth}
    \includegraphics[width=0.99\linewidth]{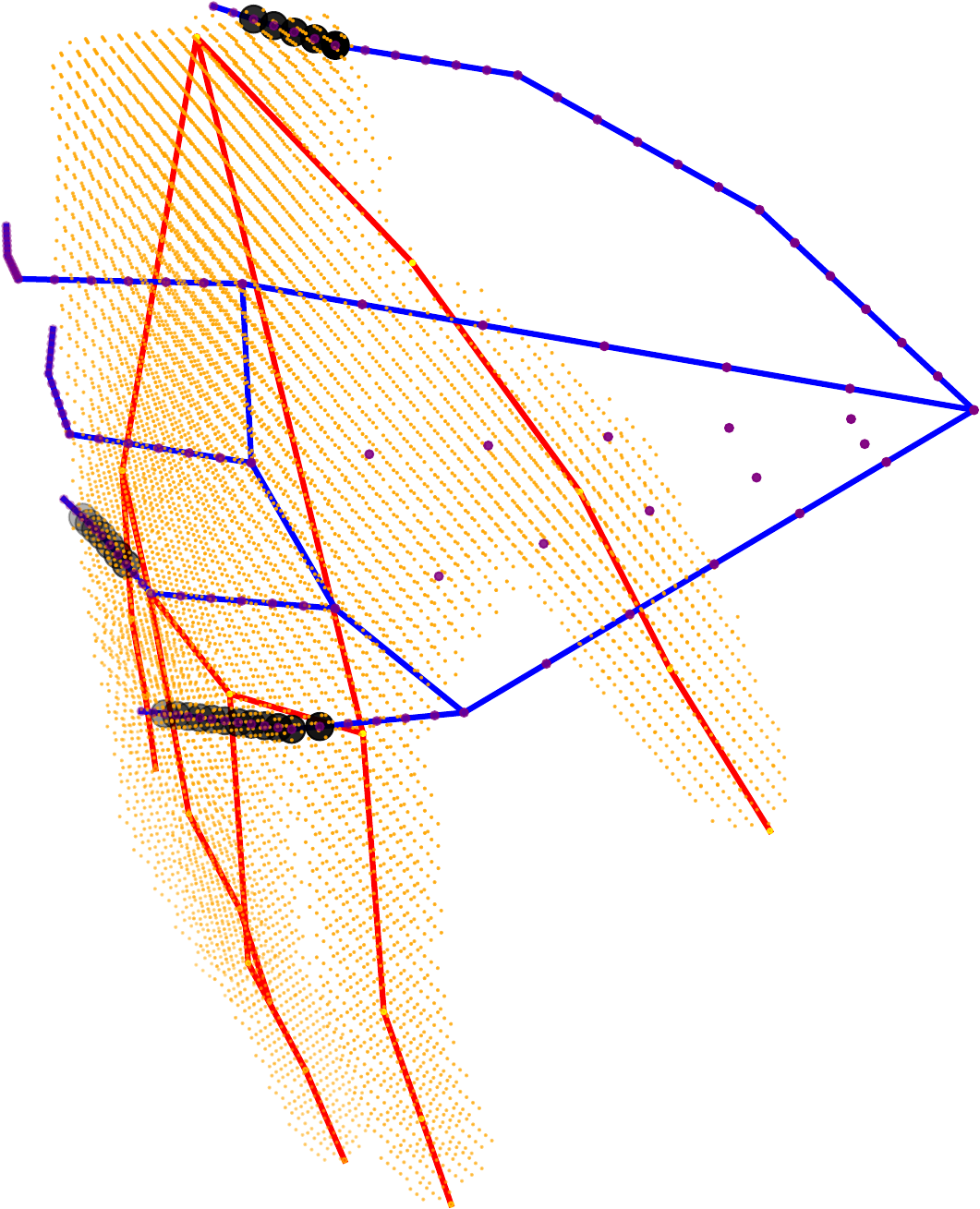}
    \caption{dense}
    \end{subfigure}
    \begin{subfigure}{0.32\linewidth}
    \includegraphics[width=0.99\linewidth]{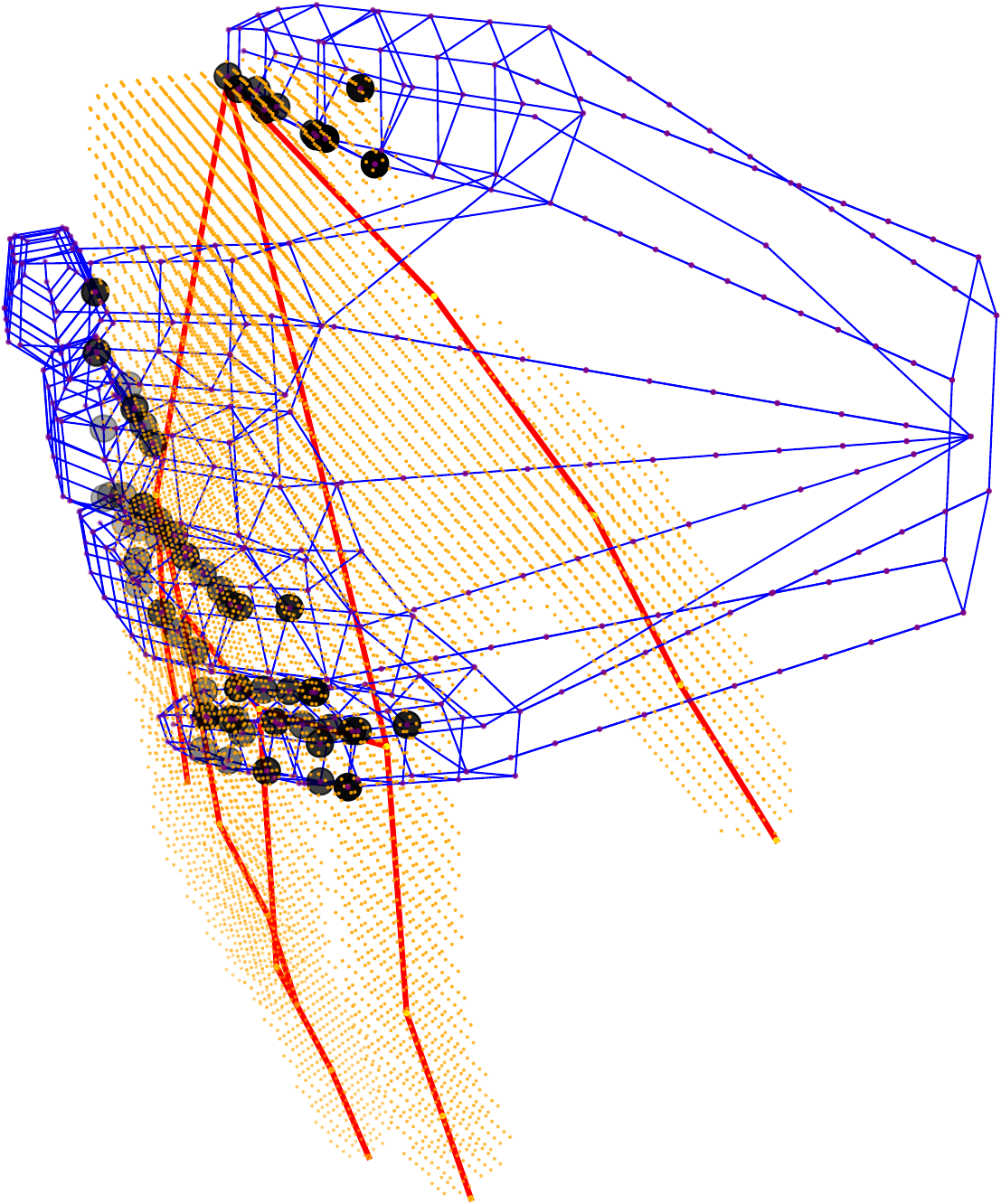}
    \caption{surface}
    \end{subfigure}
    \caption{Comparison of applying different point sets to check intersections: a) sparse skeleton, b) skeleton with additional points along the edges, c) skeleton with mesh surface points. The yellow points show the density of the right hand, and the black points highlight intersections.}
    \label{fig:loss_comparison}
\end{figure}
The pipeline in Fig.~\ref{fig:pipeline} shows hand intersections as dense point clouds estimated using the occupancy network.
However, this representation would be inefficient.
In practice, a different technique is used.
\myhighlight{For interacting hands,} let ${\bf X} \in \mathbb{R}^{3\times N}$ be a matrix of 3D hand joints stacked in a column, where the most common hand representation defines $N = 21$ points.
We distinguish joints of right and left hands by the corresponding underscore letters,~{\ie} ${\bf X}_R$ and ${\bf X}_L$.
Without loss of generality, the point set of the right hand is utilized to condition the occupancy network, and the left hand's set is employed to check point-wise intersections.
For neural networks, it is essential to work in a high-dimensional space as it enables them to capture subtle data patterns, learn features faster, and thus have better convergence.
Therefore, we exploit an encoder $\mathcal{F} : \mathbb{R}^{3\times N} \rightarrow \mathbb{F}$ that converts the point set ${\bf X}_{R}$ to a feature representation.

The goal of the intersection loss is to minimize the probabilities of the left hand's points when the occupancy network is conditioned on the right hand.
The corresponding objective is the following:
\begin{equation}
    \sum^{N}_{i=1} \biggl[\mathcal{O}\bigl({{\bf X}_L}_i, \mathcal{F}({\bf X}_R)\bigr)^2 + \alpha\mathcal{O}\bigl({\bar{{\bf X}}_R}_i, \mathcal{F}(\bar{{\bf X}}_L)\bigr)^2\biggr],
    \label{eq:intersection_loss}
\end{equation}
where ${{\bf X}_{\{L,R\}}}_i$ denotes indexing the $i$-th point of the matrix.
The parameter $\alpha \in \{0, 1\}$ is a constant that determines whether flipped hands are tested for intersection.
Correspondingly,  ${\bar{\mathbf{X}}_{R}}_i$ and ${\bar{\mathbf{X}}_{L}}_i$ denote a point set flipped along the $x$-axis ({\ie} multiplied by -1).
The square loss function is particularly effective at emphasizing points with a higher likelihood of intersection, as it progressively diminishes the impact of lower values within the range of zero to one.
Along with the squared loss, a truncated loss can also be employed, {\ie} probabilities $>$50\%.
\myhighlight{The motivation being to focus on points more likely to intersect. However, we empirically found that a non-truncated loss resolves more intersections for the same hyperparameter settings ({\eg} learning rate).}

In practice, the batched calculations are performed within the Pytorch~\cite{NEURIPS2019_9015} library, which makes the intersection loss extremely efficient, as only 21 points need to be tested.
Nevertheless, 21 joints do not span the whole hand nor its volume, and gaps between points result in a slightly less accurate intersection test.
This can be mitigated by adding virtual points along the hand edges, which introduces an additional 100 points,~{\eg} the hand skeleton contains 20 edges where 5 new points are added to each edge.
Alternatively, the whole hand mesh surface can be used for intersection verification.
However, this slows model training as the total number of points to test rises significantly.
In total, there are six options for checking hand intersections,~{\ie} testing sparse, dense, and mesh surface vertices for a single (left) hand, or additionally flipping both hands \myhighlight{to repeat the test}.
In the experiments, we provide results for each of these options.
The comparison of skeletons used for testing is demonstrated in Fig.~\ref{fig:loss_comparison}, where larger point sets reveal areas where intersections could happen.

%% file: sections/experiments.tex
\section{EXPERIMENTS}
\myhighlight{To evaluate the effect of the intersection loss and occupancy network on the pose estimation of interacting 3D hands,} we perform several sets of experiments.
Firstly, on the~\textsc{InterHand2.6M} dataset, we compare against {~\sota} methods.
\myhighlight{Secondly, on \textsc{Re:InterHand} dataset of interacting hands.} 
Finally, we train the hand pose estimation model on the SMILE dataset~\cite{ebling-etal-2018-smile}, and evaluate the accuracy on ``in the wild'' videos where the ground truth is not known.

The standard metric for 3D accuracy is the mean per-joint position error (MJPJE).
However, to measure the number of intersections from the occupancy network, we additionally exploit a ray-casting algorithm to precisely check whether points are inside a mesh to confirm the results.

\begin{figure*}
    \centering
    \includegraphics[width=0.9\linewidth]{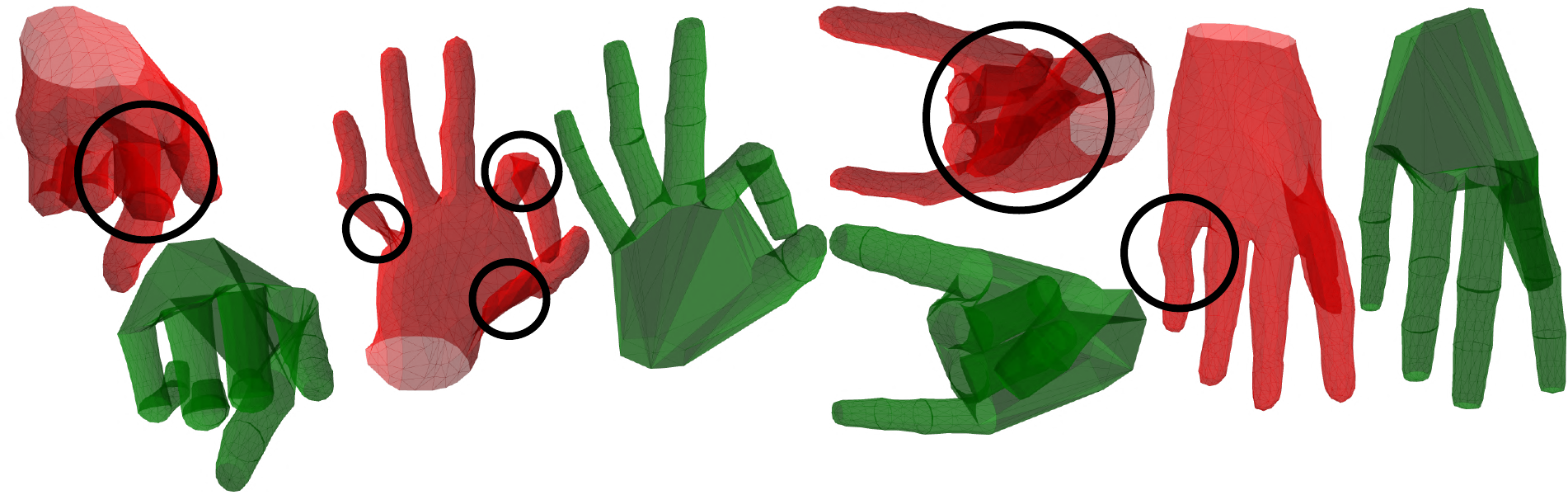}
    \vspace{-0.2cm}
    \caption{This figure shows a comparison of the failed MANO (red) and our (green) meshes fitted to the \textsc{InterHand2.6} 3D hand joints. The black circles on the MANO meshes highlight specific problems of the MANO hand's appearance, such as twisted fingers, unrealistic shape, incorrect finger orientation, etc.}
    \label{fig:mesh-comparison}
\end{figure*}
As mentioned previously, the proposed mesh is watertight, which enables us to test if an arbitrary point is inside the mesh.
We create a mesh grid of uniformly distributed 3D points that span the size of meshes.
In total, 125 thousand points are tested using the ray-casting algorithm to find mesh occupancies and mask points inside meshes for each hand.

\subsection{Training the Occupancy Network}
The occupancy network is conditioned with a single-hand skeleton that has 21 joints. Surprisingly, conditioning the occupancy network with a mesh surface instead of a sparse skeleton does not significantly increase the accuracy of the model, however, it does make training times much longer.

Random samples of points (8192 per hand) and corresponding occupancy masks are used to train the occupancy network.
During validation, all 125 thousand points are employed to compute the intersection over union rate of the ground truth and predicted occupancies, which is the primary metric for the occupancy network evaluation. 
During training, additional augmentations on the input and sampled points are employed.
We randomly rotate input skeletons and corresponding meshes by $\pm180$ degrees, and random Gaussian noise is added to the sampled mesh points.

\begin{figure}
    \centering
    \includegraphics[width=0.75\linewidth]{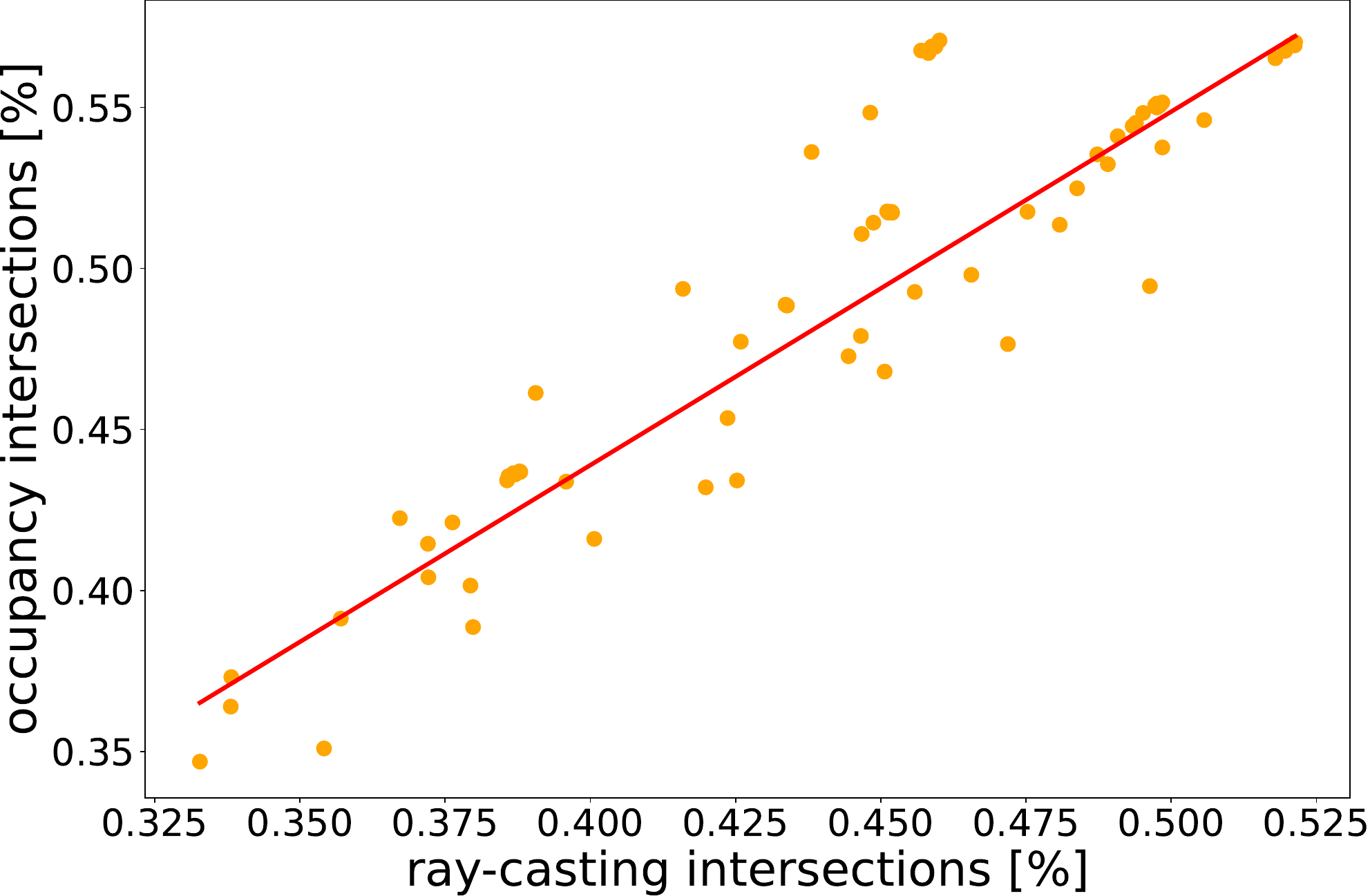}
    \vspace{-0.05cm}
    \caption{Correlation trend of per-point intersection probability found via occupancy network and ray-casting algorithm.}
    \label{fig:occ_ray_correlation}
\end{figure}

\subsection{Results on \textsc{InterHand2.6M} dataset}
\label{subsection:interhand}
The~\textsc{InterHand2.6M} dataset~\cite{Moon_2020_ECCV_InterHand2.6M} 
contains 2.6 million images of single and interacting hands, with ground truth poses found from the triangulation of 80-140 views. 
The 2D hand detections were obtained by either manual human labeling or an automatic annotation tool.
The accuracy of ~\textsc{InterHand2.6M} is notably high, as the reconstruction process leveraged a large number of different camera views.

The dataset also contains MANO meshes that were fitted to the triangulated skeletons with about a 5 mm error\footnote{\url{https://mks0601.github.io/InterHand2.6M/}}.
By exploiting the custom mesh parameterization and an inverse kinematics solver~\cite{ivashechkin_improved3d} we fit our new mesh with less than 0.5 mm error.
Having significantly more accurate meshes is needed for the occupancy network and is vital for the hand intersection test, where a small finger's shift is crucial.

\myhighlight{Where pre-trained models were available, we applied the intersection loss to the top performing methods with respect to the lowest MPJPE error on the} \textsc{InterHand2.6M} \myhighlight{dataset.}
Since not all of them provide the training code, we trained a simple linear network (MLP) that takes as input 3D interacting hand skeletons from a~{\sota} model and generates the same 3D skeletons.
The accuracy of this network matches the accuracy of the~{\sota} methods.
\myhighlight{We then took the subset of the data containing two interacting hands}, and fine-tuned the 3D-to-3D model with the intersection loss.

\begin{figure}[t]
    \centering
    \includegraphics[width=0.85\linewidth]{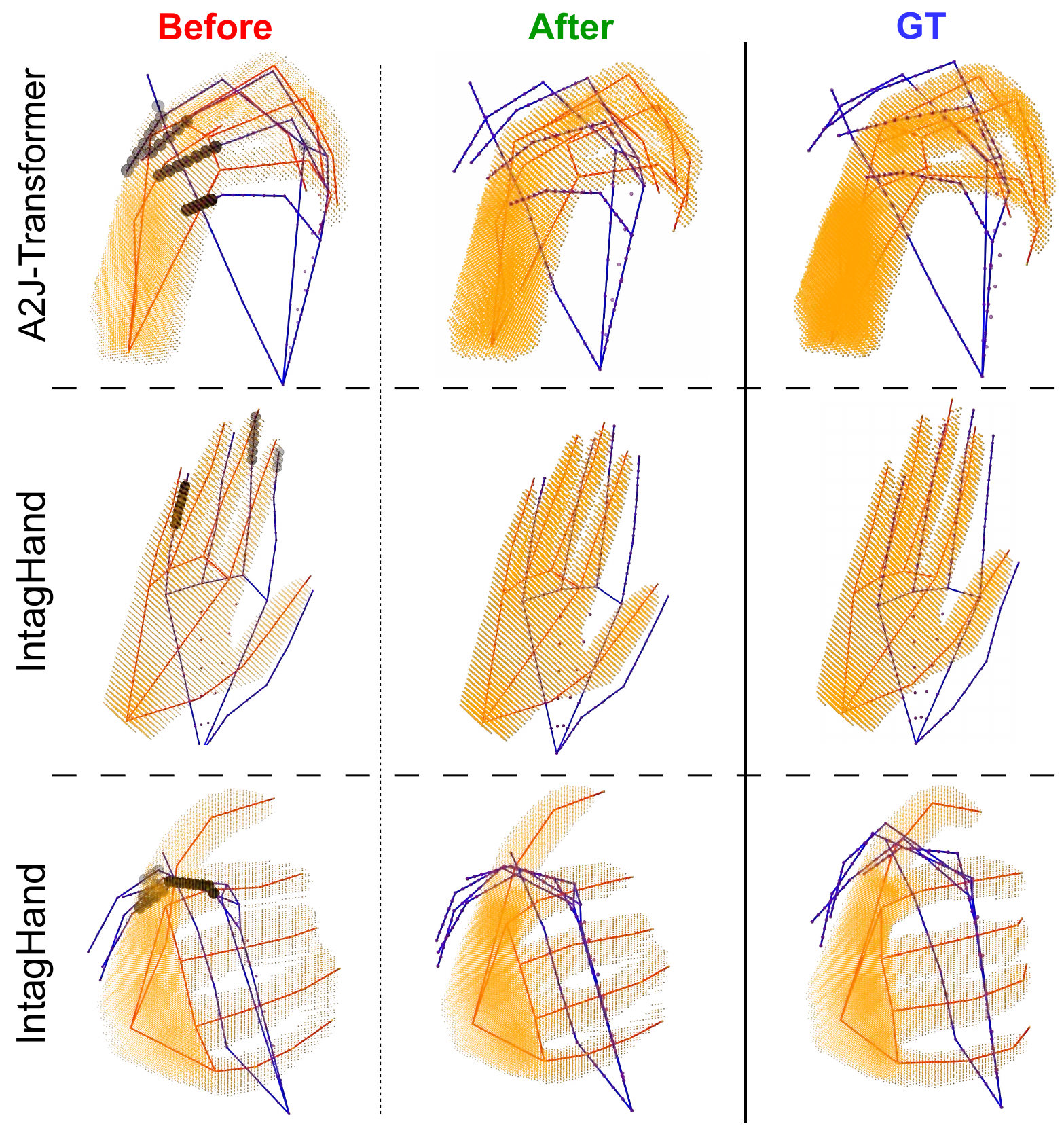}
    \vspace{-0.2cm}
    \caption{Comparison of 3D hand poses estimated by~{\sota} methods (IntagHand~\cite{Li2022intaghand} and A2J-Transformer~\cite{jiang2023a2jtransformer}) before (left) and after (middle) training with the hand intersection loss on~\textsc{InterHand2.6M}. The ground truth hands are shown in the right column. Red skeletons show right hands, blue skeletons left hands. Orange points around the right hand highlight volumetric hand density found via the occupancy network. Large black points show point-to-hand intersections. Note: there are no intersections in either the model trained with intersection loss or the ground truth.}
    \label{fig:occ_comparison}
\end{figure}

The main parameter that influences both MPJPE and intersection accuracy is the weight of the intersection loss function.
\myhighlight{If the weight is too high}, the model tries to push hands away from each other to prevent any intersections, which increases the 3D error.
Conversely, a small weight does not have any impact on the model.
We found that the optimal weight that balances lower MPJPE while minimizing the intersections is within the range $10^{-5}$ to $10^{-8}$ depending on the size of the points tested ({\eg} sparse or mesh).

\input{tables/interhand_new}

In our experiments, the fine-tuned models trained with the intersection loss consistently have a lower number of intersections and improved MPJPE over~{\sota} methods.
When comparing to the~{\sota}, we specifically selected models that achieve a similar mean per-joint position error while minimizing the overall number of intersections.

The number of intersections is found in two ways.
First, using the occupancy network with skeleton conditioning.
\myhighlight{However, to achieve a fair comparison, with less reliance on the occupancy network, we find intersections by fitting custom meshes to skeletons returned by the~{\sota}. The skeletons of new models are trained with the intersection loss}; the same way we converted the~\textsc{InterHand2.6M} 3D hands to meshes. 
The mean fitting error is lower than 0.4 mm, which provides highly precise meshes.
We have conservatively chosen a thinner volumetric shape for meshes, particularly the finger thickness, to ensure a high level of confidence in intersection detection.

Table~\ref{table:interhand} reports the number of intersections, found via the ray-casting algorithm and the occupancy network, for both~{\sota} and versions with the proposed intersection loss.
\myhighlight{We found that the best intersection loss used for fine-tuning all models employs a sparse set of points of a single hand with a non-truncated kernel.}
\myhighlight{In the experiments, these settings resulted in the best reduction of intersections and MPJPE.}
\myhighlight{The proposed CNN model provides excellent MPJPE accuracy having an 11.4 mm error compared to the A2J~\cite{Xiong_a2j} method (11.2 mm), or the third place Keypoint Transformer~\cite{Hampali_2022_CVPR_Kypt_Trans} (14.7 mm). } 
In Table~\ref{table:interhand}, the percentage decrease in the number of intersections found via the ray-casting algorithm correlates with intersections detected with the occupancy network, \myhighlight{see Fig.~\ref{fig:occ_ray_correlation}}.
Therefore, it additionally confirms the accuracy of the occupancy model.


\myhighlight{Models including the intersection loss achieve }significantly fewer hand-to-hand intersections than~{\sota}.
For example, Fig.~\ref{fig:loss_comparison} shows a comparison of interacting hands of IntagHand and  A2J-Transformer models against their versions trained with the intersection loss.

Removing all intersections is challenging for several reasons:
Firstly, a higher weight for the intersection loss may lead to significant deterioration in 3D accuracy of hand estimator model.
Secondly, there is not enough data for hands in close interaction where possible intersections could occur. Such situations correspond to approximately 17\% percent of the~\textsc{InterHand2.6M} data.
Thirdly, error propagation from the occupancy network, which in experiments on the validation set has around 80\% of intersection over the union (IoU) accuracy.
Finally, imprecision of the ground truth data that also contains intersections, as the 3D ground skeletons are often extracted with triangulation algorithms. 

For the same reasons, lowering the mean per-joint position error is difficult.
Since we use an unsupervised intersection loss, there is no guidance to the model on how it should fix the intersections,{~\ie} the intersection loss function only tells the model where it cannot position the hand joints.
During training, the best model was selected using the minimum error on the validation set in terms of mean per-joint accuracy (not the minimum amount of intersections).

\begin{figure*}
    \centering
    \includegraphics[width=0.85\linewidth]{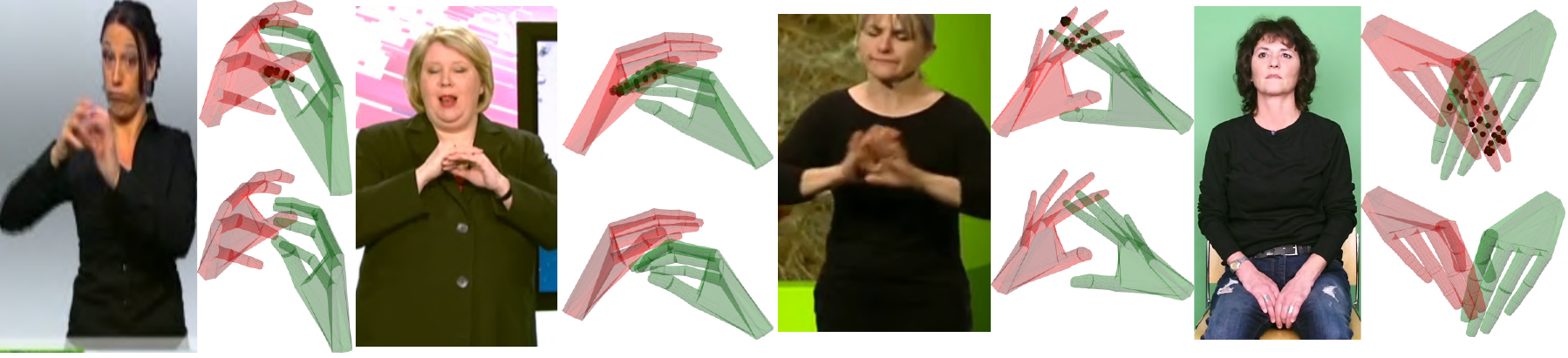}
    \caption{Comparison of 3D hand estimation models after applying the intersections loss (bottom) and without (top) on images taken from various sign language datasets not used in training (left to right): RWTH-Phoenix Weather 2014~\cite{koller15:cslr}, German sign language DSGS\cite{li-etal-2019-findings}, BBC-Oxford British Sign Language~\cite{Albanie2020bsl1k,Albanie2021bobsl}, and SMILE Swiss sign language dataset~\cite{ebling-etal-2018-smile}. The black points highlight point-wise intersections. The model trained with intersection loss produced 3D hands completely free of intersections.
    Consequently, the bottom meshes are the refined version of the custom mesh parameterization.}
    \label{fig:in_the_wild_qualitative}
\end{figure*}

\subsection{Results on the \textsc{Re:InterHand} dataset}
\vspace{-0.1cm}
The \textsc{Re:InterHand} dataset~\cite{moon2023reinterhand} of Moon~{\etal} provides a large collection of synthetically generated realistic images of interacting hands.
In our experiments, we used egocentric viewpoints where 8 captures are reserved for training (402200 frames) and the remaining 2 captures for testing (90380 frames). Since the dataset is recent, we have not found any released competitor models. Therefore, we used a CNN model to predict the two 3D hands directly from the image. Then, we fine-tuned the corresponding model with intersection loss (testing dense points of a single hand). A $10^{-4}$ weight on intersection loss decreased the error by 0.15\% and intersections (found via ray-casting) by 27\% compared to the model trained without intersection loss. Similarly, a $10^{-5}$ weight decreased the error by 3.1\% and the intersections by 6.9\%, and a $10^{-6}$ weight by 3.8\% and 3.6\% (respectively). Therefore, experiments on this dataset confirm that the weight of the intersection loss provides a good trade-off that minimizes error and hand intersections.

\subsection{In The Wild Evaluation}
It is important to evaluate the hand pose estimator and its performance ``in the wild''.
This means randomly selected videos where the quality of interacting hands is crucial for understanding,~{\eg} sign language.
However, such videos do not have the corresponding 3D ground truth reconstruction.
Therefore, the evaluation metrics consist of qualitative results as well as the number of intersections determined via the ray-casting algorithm, as shown on~\textsc {InterHand2.6M}.


We designed a multi-layer perception (MLP) to uplift right and left 3D hands with hand-to-hand offset from 2D keypoint detections (obtained with MediaPipe~\cite{mediapipe}).
The MLP has three submodules that separately predict joint angles of the hands, the bone lengths for both hands, and the relative offset between the two hands, as in~\cite{ivashechkin_improved3d}.
The total size of the network is 9.5 million parameters.
Without loss of generality, the MLP predicts two right hands and flips the left hand in the $x$-axis.
The prediction of angles enables it to directly find the corresponding meshes with the proposed mesh parameterization.

For training the MLP, we used a version of the SMILE~\cite{ebling-etal-2018-smile} dataset that has the 3D hands triangulated from three calibrated cameras.
This makes the ground truth reconstruction significantly less reliable (in terms of interacting hands accuracy) compared to the~\textsc {InterHand2.6M} dataset.
However, it provides a greater challenge to remove intersections and lower the MPJPE error.
We opted for the SMILE dataset for several key reasons. Firstly, it enables us to show the versatility of the intersection loss when applied to an alternative dataset. Secondly, we can evaluate the effectiveness of the intersection loss in scenarios where ground truth accuracy for interacting hands is less than ideal. Lastly, as a sign language dataset, it provides a significantly larger and diverse set of real-world interacting hand data, enhancing the depth and practicality of the evaluation.

We first train a baseline network that is supervised with the ground truth 3D skeletons.
Following the training of the baseline model, it was fine-tuned using interacting hands, and the cloned models were fine-tuned with different types of the intersection loss,~{\ie} testing single/both hands or a sparse/dense/mesh surface point sets.
All models were given the same number of epochs to converge.

\input{tables/in_the_wild}

\myhighlight{Table~\ref{table:in_the_wild_occs} compares the number of hand intersections (found via occupancy network) for the baseline and models trained with intersection loss.}
Similarly to the experiments on the~\textsc {InterHand2.6M} dataset, we selected models that have around the same error as the baseline, but the least number of intersections.
Table~\ref{table:in_the_wild_sign_datasets} shows the number of intersections found with the ray-casting algorithm on four ``in the wild'' sign-language datasets.
In each dataset, we selected 250000 random interacting hand pairs, except for the Phoenix dataset~\cite{koller15:cslr}, which has around 120000 pairs; and is the only dataset where models with the intersection loss do not fully outperform the baseline model. 
The smaller quantity of test data could be the reason for this. 
From Table~\ref{table:in_the_wild_occs} and~\ref{table:in_the_wild_sign_datasets} the results suggest that models with the intersection loss resolve a substantial amount of intersections.
Moreover, the decrease in the number of intersections found via the occupancy network and ray-casting algorithm (averaged over datasets) correlates, similar to the results on the~\textsc{InterHand2.6M} dataset, which again confirms the accuracy of the occupancy network.
In the majority of cases, the model trained using the dense points intersection loss yields the fewest intersections.
Testing mesh surface vertices results in a greater reduction of intersections compared to a sparse points set.
Nevertheless, with a significantly larger distribution of points in space, the model's performance unexpectedly declines.
Testing both hands shows (on average) a further reduction in intersections compared to a single hand. 
The qualitative evaluation of the baseline and models with intersection loss is shown in Fig.~\ref{fig:in_the_wild_qualitative}, where hands returned by the network with the intersection loss are more physically plausible.

\input{tables/time_difference}
\subsection{Time complexity}
To investigate the impact of testing different point sets in the intersection loss against the model training time,~{\eg} sparse/dense/surface or the number of hands, we measured the amount of iterations executed per second ({\ie} model forward pass, loss computation, backpropagation, and parameter update) for a batch size of 256 interacting hands. 
Table~\ref{table:time} shows the time comparison for each setting, where testing both hands in the intersection loss accounts for around 50\% of computational time.
\myhighlight{The computational complexity of the intersection loss increases with the number of points tested.}
The fastest option (sparse, single hand) is more than three times faster than the slowest (surface points with two hands), and around 36\% slower than training a model without an intersection loss.
Note, the values in table~\ref{table:time} are obtained by running occupancy on all interacting hands.
\myhighlight{However, the additional speed-up is gained by filtering the hand pairs with a 3D bounding box intersection test.}
The measurements were done on the 11th Gen Intel Core i9 Ubuntu machine with an NVIDIA GeForce RTX 3090 GPU.

Considering that the 3D ground truth of the SMILE dataset is not very accurate, it is noteworthy that the models with intersection loss have managed to decrease intersections while lowering or maintaining the MPJPE accuracy.
Testing dense points of a single hand provides a good trade-off maintaining training speed while ensuring minimal intersections. 

\subsection{Noise Influence on Intersection Loss}
\begin{figure}
    \centering
    \includegraphics[width=0.8\linewidth]{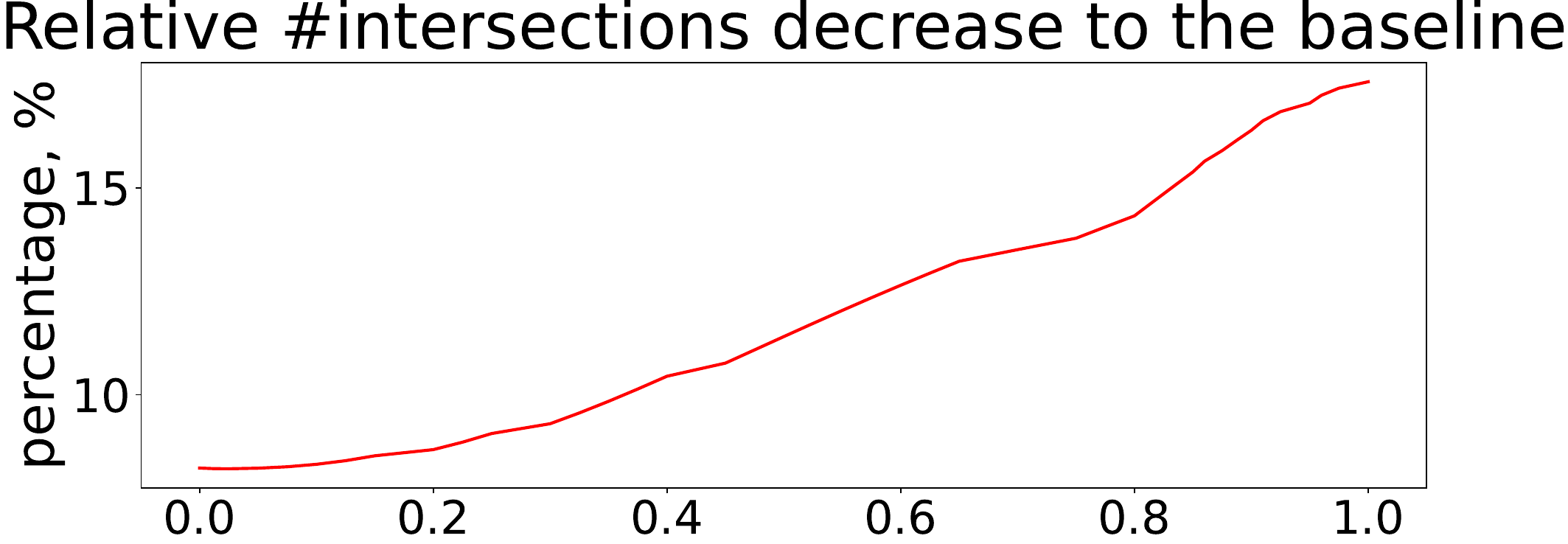}
    \includegraphics[width=0.8\linewidth]{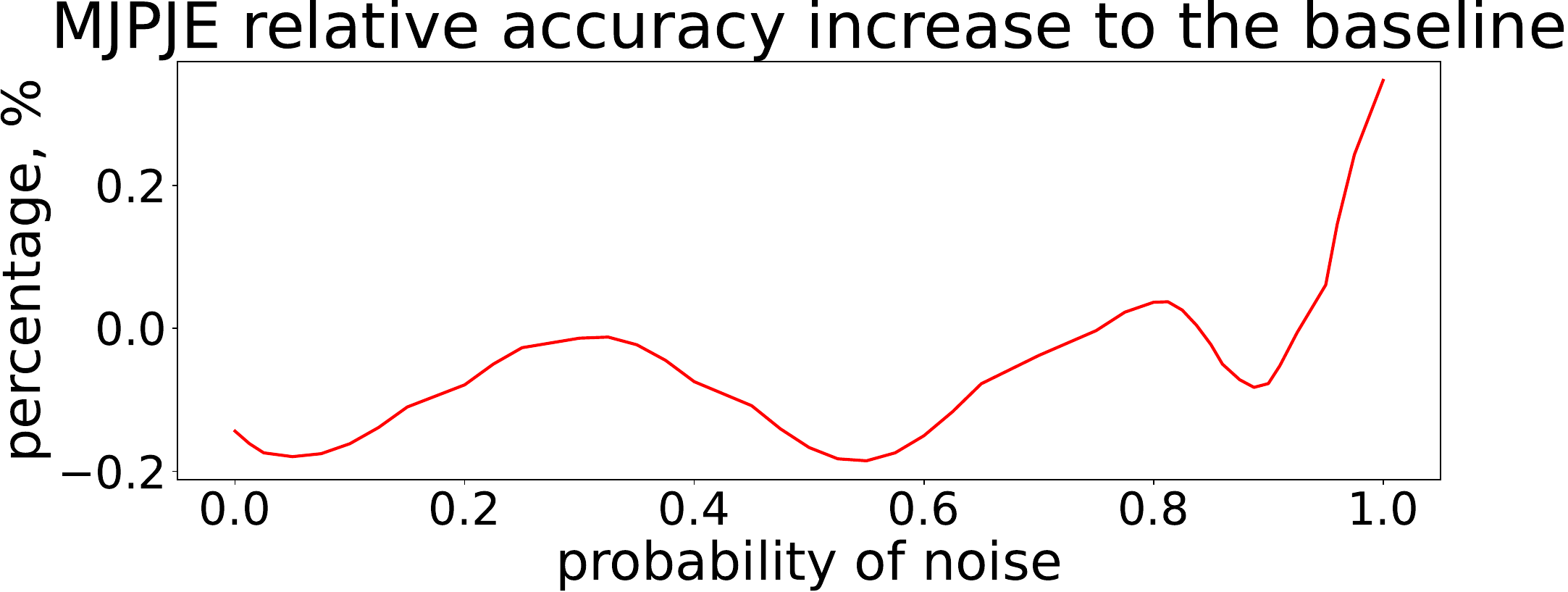}
    \caption{The plots show the impact of artificially noising 3D training data on mean per-joint position error (MJPJE) and the number of intersections found with the occupancy network. For each noise probability, we compared two models -- with and without intersection loss. The top graph demonstrates a decrease in the intersections. The bottom shows an increase in 3D accuracy for the model with intersection loss.}
    \label{fig:noise_experiment}
\end{figure}

\vspace{-0.1cm}
The intersection can provide additional cues to the 3D uplift model by providing information where hand joints cannot be located to avoid hand intersection.
To confirm this hypothesis, we designed an experiment, where during training, the ground truth 3D points of highly accurate hands from the~\textsc{InterHand2.6M} dataset were artificially noised (both hands slightly rotated) with a range of probabilities from 0 to 1.
The objective is to demonstrate that the model with intersection loss performs better than its version without, as the model with intersection loss can avoid intersections and thus should be closer to the ground truth.
Fig.~\ref{fig:noise_experiment} shows that with a higher probability of noise, the model with intersection loss linearly outperforms the baseline model (the number of intersections and is slightly better in MPJPE for noise probability higher than 0.8).
This empirically proves the initial assumption, since the figure suggests that the baseline model has no perception of hand intersections, and higher noise leads to a less accurate estimate.
The model with intersection loss results in better prediction.

\subsection{Implementation details}
The intersection loss requires a pre-trained occupancy network 
which consists of a PointNet encoder~\cite{pointnet} with residual blocks that encode input 3D skeletons into a feature space and a decoder with conditional batch normalization that transforms the latent space into logits.
In total, the occupancy network has 7 million parameters.

\myhighlight{A CNN network for predicting the hands consists of a ResNet-50~\cite{resnet} model that returns image features, and an MLP with a couple of fully connected linear layers (in total around 2m parameters) that regresses a 3D hand pose.
The predicted hands from the MLP are used in the intersection loss with the occupancy network.}

\subsection{Limitations}
As was mentioned in the section~\ref{subsection:interhand} describing the results on the \textsc{InterHand2.6M} dataset, the main limitation of our approach is that not all intersections are resolved. 
Primarily, this is caused by the weight of intersection loss in the model training, where a high weight leads to solving more intersections but deteriorates the 3D accuracy.
A smaller weight, on the other hand, fixes fewer intersections but lowers the 3D error.

%% file: tables/interhand_new.tex

\begin{table}[ht]
\setlength{\tabcolsep}{5pt}
\renewcommand{\arraystretch}{0.95}
\centering
\caption{Comparison of~{\sota} models and their updated versions (``+Ours'') trained with intersection loss in terms of a number of intersections.
The second column indicates the amount of intersections found via the ray-casting algorithm (``Ray-C.''), and the last column shows the quantity of intersections determined by the occupancy network (``Occ.''). The ``\%'' row reports a percentage decrease in the amount of intersections for the proposed models. The total number of tested hand pairs is around 220000. CNN stands for our baseline model.}
\vspace{-0.15cm}

\begin{tabular}{lrr||lrr}
\rowcolor{tablegray}\hline
\rowcolor{tablegray}
\multicolumn{1}{c}{Method} & \multicolumn{1}{c}{Ray-C.} & \multicolumn{1}{c||}{Occ.} & \multicolumn{1}{c}{Method} & \multicolumn{1}{c}{Ray-C.} & \multicolumn{1}{c}{Occ.} \\
Xiong~\cite{Xiong_a2j} & 376111 & 20771 & Moon~\cite{Moon_2020_ECCV_InterHand2.6M} &418079 & 22233\\
Xiong+Ours & {\bf 306478} & {\bf 17835} & Moon+Ours & {\bf 295261} & {\bf 16269} \\
\% & 18.51 & 14.13  & \% &   29.38 & 26.82 \\ \hline
Li~\cite{Xiong_a2j} & 583062 & 32253 & Zhang~\cite{Zhang2021twohand} & 592685 & 32240  \\
Li+Ours & {\bf 372534} & {\bf 20286}  & Zhang+Ours & {\bf 336591} & {\bf 18843}  \\
\% & 36.11 &  37.10  &\%  &  43.21 &  41.55\\ \hline
Hampali~\cite{Hampali_2022_CVPR_Kypt_Trans} & 452408 & 25765 & CNN & 490212 &  25904 \\
Hampali+Ours & {\bf 325803} & {\bf 19827} & CNN+Ours  & {\bf 394345} & {\bf 22184}  \\
\% & 27.98 & 23.05 & \% &  19.56 & 14.36 \\\hline
\end{tabular}
\label{table:interhand}
\end{table}

%% file: tables/in_the_wild.tex
\begin{table}[ht]
\centering
\caption{Percentage decrease in intersections (found with an occupancy network) for the baseline model vs models trained with intersection loss. The rows show whether single or both hands were employed.
The columns indicate the density of points. The results correspond to the test set (in total 734120 pairs) of the SMILE dataset.}
\vspace{-0.15cm}
\begin{tabular}{lrrr}
\rowcolor{tablegray}\hline
\multicolumn{4}{c}{Number of intersections decrease in \%} \\
\rowcolor{tablegray}
& \multicolumn{1}{c}{sparse} & \multicolumn{1}{c}{dense} & \multicolumn{1}{c}{mesh} \\ \hline
\multicolumn{1}{c}{single} &8.189 & 16.346 & 9.444  \\
\multicolumn{1}{c}{both} & 5.361 & {\bf 18.727} & 7.167 \\\hline

\end{tabular}
\label{table:in_the_wild_occs}
\end{table}


\begin{table}[ht]
\renewcommand{\arraystretch}{0.95}
\caption{
Percentage decrease in the number of intersections found via the ray-casting algorithm comparing the baseline network and models trained with different types of intersection loss on four sign-language datasets.
The rows indicate whether single or both hands were tested, and the columns show the density of the point sets.}
\vspace{-0.15cm}
\begin{subtable}[c]{0.24\textwidth}
\centering
\begin{tabular}{p{0.8cm}p{0.6cm}p{0.6cm}p{0.6cm}}
\rowcolor{tablegray}\hline
\multicolumn{4}{c}{BOBSL~\cite{Albanie2020bsl1k,Albanie2021bobsl}} \\
\rowcolor{tablegray}
& \multicolumn{1}{c}{S} & \multicolumn{1}{c}{D} & \multicolumn{1}{c}{M} \\
\multicolumn{1}{c}{single} & 3.94 & 16.30 & 8.88 \\
\multicolumn{1}{c}{both} &5.18 & {\bf 16.59} & 9.67 \\
\end{tabular}
\end{subtable}
\hfill
\begin{subtable}[c]{0.24\textwidth}
\centering
\begin{tabular}{p{0.8cm}p{0.6cm}p{0.6cm}p{0.6cm}}
\rowcolor{tablegray}\hline
\multicolumn{4}{c}{DSGS~\cite{li-etal-2019-findings}} \\
\rowcolor{tablegray}
& \multicolumn{1}{c}{S} & \multicolumn{1}{c}{D} & \multicolumn{1}{c}{M} \\
\multicolumn{1}{c}{single} & \phantom{1}4.04 & 11.47 & \phantom{1}7.46 \\
\multicolumn{1}{c}{both} &\phantom{1}3.97 & {\bf 14.97} & \phantom{1}5.41 \\
\end{tabular}
\end{subtable}
\hfill
\begin{subtable}[c]{0.24\textwidth}
\centering
\begin{tabular}{p{0.8cm}p{0.6cm}p{0.6cm}p{0.6cm}}
\rowcolor{tablegray}\hline
\multicolumn{4}{c}{Phoenix~\cite{koller15:cslr}
} \\
\rowcolor{tablegray}
& \multicolumn{1}{c}{S} & \multicolumn{1}{c}{D} & \multicolumn{1}{c}{M} \\ 
\multicolumn{1}{c}{single} & 2.53 & \phantom{1}3.59 & 2.76 \\
\multicolumn{1}{c}{both} &0.63 & \phantom{1}{\bf 7.87} & 2.06 \\ 
\end{tabular}
\end{subtable}
\hfill
\begin{subtable}[c]{0.24\textwidth}
\centering
\begin{tabular}{p{0.8cm}p{0.6cm}p{0.6cm}p{0.6cm}}
\rowcolor{tablegray}\hline
\multicolumn{4}{c}{SMILE~\cite{ebling-etal-2018-smile}} \\
\rowcolor{tablegray}
& \multicolumn{1}{c}{S} & \multicolumn{1}{c}{D} & \multicolumn{1}{c}{M} \\
\multicolumn{1}{c}{single} & \phantom{1}8.78 & 20.16 & 12.74 \\
\multicolumn{1}{c}{both}& 10.84 & {\bf 31.20} & 15.52 \\ 
\end{tabular}
\end{subtable}
\centering
\begin{tabular}{p{0.8cm}p{2.05cm}p{2.05cm}p{2.05cm}}
\rowcolor{tablegray}\hline
\multicolumn{4}{c}{Average over datasets} \\
\rowcolor{tablegray}
& \phantom{000000}S & \phantom{0000000}D & \phantom{000000}M \\
\multicolumn{1}{c}{single} & \phantom{00000}4.82 & \phantom{00000}12.88 & \phantom{00000}7.96 \\
\multicolumn{1}{c}{both} & \phantom{00000}5.15 & \phantom{00000}{\bf 17.66} & \phantom{00000}8.16 \\ \hline
\end{tabular}
\label{table:in_the_wild_sign_datasets}
\end{table}

%% file: tables/time_difference.tex
\begin{table}[ht]
\centering
\caption{
Comparison of training times across models with varying types of intersection loss.
Rows specify the test type and rows the number of hands tested. The model without intersection loss is in the bottom row.
}
\vspace{-0.15cm}

\begin{tabular}{lrrr}
\rowcolor{tablegray}\hline
\multicolumn{4}{c}{Iterations per second} \\
\rowcolor{tablegray}
& \multicolumn{1}{c}{sparse} & \multicolumn{1}{c}{dense} & \multicolumn{1}{c}{mesh} \\\hline
\multicolumn{1}{c}{single} &  13.23 & 11.72 & 6.62 \\
\multicolumn{1}{c}{both} &  9.88 & 8.22 & 4.19 \\\cdashline{1-4}
\multicolumn{1}{c}{no intersection loss} & \multicolumn{3}{c}{20.85}\\
\hline
\end{tabular}
\label{table:time}
\end{table}

%% file: sections/conclusions.tex
\section{CONCLUSIONS}
\vspace{-0.1cm}
This paper presents an intersection loss function for interacting 3D hand estimation to introduce physical constraints on hand estimation.
By exploiting an occupancy network conditioned on a 3D skeleton, the hand volume is represented as a continuous manifold, where for an arbitrary 3D point, the occupancy model represents the likelihood of a hand intersection.
With an extensive ablation study, we investigated the impact of testing different point sets for hand intersection on the model's accuracy and training speed.
Additionally, we propose a custom hand mesh parameterization that overcomes some of the limitations of the MANO model, such as lower complexity, access to underlying 3D skeleton, watertightness, etc.
The custom mesh serves a versatile purpose, enabling fast mesh generation essential for our occupancy network or mesh rendering, while providing a refined version tailored for mesh visualization.

The experiments on the benchmark~\textsc{InterHand2.6M} dataset improve~{\sota} models in both mean per-joint position error and significantly decrease the number of intersections.
The cross-validation on the \textsc{Re:InterHand} and ``in the wild'' videos of sign language confirms a reduction of hand intersections both quantitatively and qualitatively, while maintaining the same 3D accuracy even though the hand estimator was trained on a very noisy dataset.

%% file: arxiv-version.bbl
\begin{thebibliography}{10}\itemsep=-1pt

\bibitem{Albanie2020bsl1k}
S.~Albanie, G.~Varol, L.~Momeni, T.~Afouras, J.~S. Chung, N.~Fox, and A.~Zisserman.
\newblock {BSL-1K}: {S}caling up co-articulated sign language recognition using mouthing cues.
\newblock In {\em European Conference on Computer Vision}, 2020.

\bibitem{Albanie2021bobsl}
S.~Albanie, G.~Varol, L.~Momeni, H.~Bull, T.~Afouras, H.~Chowdhury, N.~Fox, B.~Woll, R.~Cooper, A.~McParland, and A.~Zisserman.
\newblock {BOBSL}: {BBC}-{O}xford {B}ritish {S}ign {L}anguage {D}ataset.
\newblock 2021.

\bibitem{ballan_motion}
L.~Ballan, A.~Taneja, J.~Gall, L.~Van~Gool, and M.~Pollefeys.
\newblock Motion capture of hands in action using discriminative salient points.
\newblock In {\em Computer Vision -- ECCV 2012}, pages 640--653, Berlin, Heidelberg, 2012. Springer Berlin Heidelberg.

\bibitem{Zhang2021InteractingT3}
Z.~Baowen, W.~Yangang, D.~Xiaoming, Z.~Yinda, T.~Ping, M.~Cuixia, and W.~Hongan.
\newblock Interacting two-hand 3d pose and shape reconstruction from single color image.
\newblock {\em 2021 IEEE/CVF International Conference on Computer Vision (ICCV)}, pages 11334--11343, 2021.

\bibitem{pointnet}
R.~Q. Charles, H.~Su, M.~Kaichun, and L.~J. Guibas.
\newblock Pointnet: Deep learning on point sets for 3d classification and segmentation.
\newblock In {\em 2017 IEEE Conference on Computer Vision and Pattern Recognition (CVPR)}, pages 77--85, 2017.

\bibitem{Chen2018GeneratingRT}
L.~Chen, S.-Y. Lin, Y.~Xie, H.~Tang, Y.~Xue, X.~Xie, Y.-Y. Lin, and W.~Fan.
\newblock Generating realistic training images based on tonality-alignment generative adversarial networks for hand pose estimation.
\newblock {\em ArXiv}, abs/1811.09916, 2018.

\bibitem{ebling-etal-2018-smile}
S.~Ebling, N.~C. Camg{\"o}z, P.~Boyes~Braem, K.~Tissi, S.~Sidler-Miserez, S.~Stoll, S.~Hadfield, T.~Haug, R.~Bowden, S.~Tornay, M.~Razavi, and M.~Magimai-Doss.
\newblock {SMILE} {S}wiss {G}erman sign language dataset.
\newblock In {\em Proceedings of the Eleventh International Conference on Language Resources and Evaluation ({LREC} 2018)}, Miyazaki, Japan, May 2018. European Language Resources Association (ELRA).

\bibitem{fan2021digit}
Z.~Fan, A.~Spurr, M.~Kocabas, S.~Tang, M.~Black, and O.~Hilliges.
\newblock Learning to disambiguate strongly interacting hands via probabilistic per-pixel part segmentation.
\newblock In {\em International Conference on 3D Vision (3DV)}, 2021.

\bibitem{Hampali_2022_CVPR_Kypt_Trans}
S.~Hampali, S.~D. Sarkar, M.~Rad, and V.~Lepetit.
\newblock Keypoint transformer: Solving joint identification in challenging hands and object interactions for accurate 3d pose estimation.
\newblock In {\em IEEE Computer Vision and Pattern Recognition Conference}, 2022.

\bibitem{resnet}
K.~He, X.~Zhang, S.~Ren, and J.~Sun.
\newblock Deep residual learning for image recognition.
\newblock In {\em 2016 IEEE Conference on Computer Vision and Pattern Recognition (CVPR)}, pages 770--778, 2016.

\bibitem{ivashechkin_improved3d}
M.~Ivashechkin, O.~Mendez, and R.~Bowden.
\newblock Improving 3d pose estimation for sign language.
\newblock In {\em 2023 IEEE International Conference on Acoustics, Speech, and Signal Processing Workshops (ICASSPW)}, pages 1--5, 2023.

\bibitem{jiang2023a2jtransformer}
C.~Jiang, Y.~Xiao, C.~Wu, M.~Zhang, J.~Zheng, Z.~Cao, and J.~T. Zhou.
\newblock A2j-transformer: Anchor-to-joint transformer network for 3d interacting hand pose estimation from a single rgb image, 2023.

\bibitem{karunratanakul2021halo}
K.~Karunratanakul, A.~Spurr, Z.~Fan, O.~Hilliges, and S.~Tang.
\newblock A skeleton-driven neural occupancy representation for articulated hands.
\newblock In {\em International Conference on 3D Vision (3DV)}, 2021.

\bibitem{koller15:cslr}
O.~Koller, J.~Forster, and H.~Ney.
\newblock Continuous sign language recognition: Towards large vocabulary statistical recognition systems handling multiple signers.
\newblock {\em Computer Vision and Image Understanding}, 141:108--125, Dec. 2015.

\bibitem{Li2022intaghand}
M.~Li, L.~An, H.~Zhang, L.~Wu, F.~Chen, T.~Yu, and Y.~Liu.
\newblock Interacting attention graph for single image two-hand reconstruction.
\newblock In {\em IEEE/CVF Conf. on Computer Vision and Pattern Recognition (CVPR)}, June 2022.

\bibitem{li-etal-2019-findings}
X.~Li, P.~Michel, A.~Anastasopoulos, Y.~Belinkov, N.~Durrani, O.~Firat, P.~Koehn, G.~Neubig, J.~Pino, and H.~Sajjad.
\newblock Findings of the first shared task on machine translation robustness.
\newblock In {\em Proceedings of the Fourth Conference on Machine Translation (Volume 2: Shared Task Papers, Day 1)}, pages 91--102, Florence, Italy, Aug. 2019. Association for Computational Linguistics.

\bibitem{mediapipe}
C.~Lugaresi, J.~Tang, H.~Nash, C.~McClanahan, E.~Uboweja, M.~Hays, F.~Zhang, C.~Chang, M.~G. Yong, J.~Lee, W.~Chang, W.~Hua, M.~Georg, and M.~Grundmann.
\newblock Mediapipe: {A} framework for building perception pipelines.
\newblock {\em CoRR}, abs/1906.08172, 2019.

\bibitem{meng2022hdr}
H.~Meng, S.~Jin, W.~Liu, C.~Qian, M.~Lin, W.~Ouyang, and P.~Luo.
\newblock 3d interacting hand pose estimation by hand de-occlusion and removal.
\newblock October 2022.

\bibitem{OccupancyNetworks}
L.~Mescheder, M.~Oechsle, M.~Niemeyer, S.~Nowozin, and A.~Geiger.
\newblock Occupancy networks: Learning 3d reconstruction in function space.
\newblock In {\em Proceedings IEEE Conf. on Computer Vision and Pattern Recognition (CVPR)}, 2019.

\bibitem{moon2023reinterhand}
G.~Moon, S.~Saito, W.~Xu, R.~Joshi, J.~Buffalini, H.~Bellan, N.~Rosen, J.~Richardson, M.~Mallorie, P.~Bree, T.~Simon, B.~Peng, S.~Garg, K.~McPhail, and T.~Shiratori.
\newblock A dataset of relighted {3D} interacting hands.
\newblock In {\em NeurIPS Track on Datasets and Benchmarks}, 2023.

\bibitem{Moon_2020_ECCV_InterHand2.6M}
G.~Moon, S.-I. Yu, H.~Wen, T.~Shiratori, and K.~M. Lee.
\newblock Interhand2.6m: A dataset and baseline for 3d interacting hand pose estimation from a single rgb image.
\newblock In {\em European Conference on Computer Vision (ECCV)}, 2020.

\bibitem{Mueller_realtime}
F.~Mueller, M.~Davis, F.~Bernard, O.~Sotnychenko, M.~Verschoor, M.~A. Otaduy, D.~Casas, and C.~Theobalt.
\newblock Real-time pose and shape reconstruction of two interacting hands with a single depth camera.
\newblock {\em ACM Trans. Graph.}, 38(4), jul 2019.

\bibitem{Oikonomidis_tracking}
I.~Oikonomidis, N.~Kyriazis, and A.~A. Argyros.
\newblock Tracking the articulated motion of two strongly interacting hands.
\newblock In {\em 2012 IEEE Conference on Computer Vision and Pattern Recognition}, pages 1862--1869, 2012.

\bibitem{NEURIPS2019_9015}
A.~Paszke, S.~Gross, F.~Massa, A.~Lerer, J.~Bradbury, G.~Chanan, T.~Killeen, Z.~Lin, N.~Gimelshein, L.~Antiga, A.~Desmaison, A.~Kopf, E.~Yang, Z.~DeVito, M.~Raison, A.~Tejani, S.~Chilamkurthy, B.~Steiner, L.~Fang, J.~Bai, and S.~Chintala.
\newblock Pytorch: An imperative style, high-performance deep learning library.
\newblock In H.~Wallach, H.~Larochelle, A.~Beygelzimer, F.~d\textquotesingle Alch\'{e}-Buc, E.~Fox, and R.~Garnett, editors, {\em Advances in Neural Information Processing Systems 32}, pages 8024--8035. Curran Associates, Inc., 2019.

\bibitem{MANO:SIGGRAPHASIA:2017}
J.~Romero, D.~Tzionas, and M.~J. Black.
\newblock Embodied hands: Modeling and capturing hands and bodies together.
\newblock {\em ACM Transactions on Graphics, (Proc. SIGGRAPH Asia)}, 36(6), Nov. 2017.

\bibitem{rong2021ihmr}
Y.~Rong, J.~Wang, Z.~Liu, and C.~C. Loy.
\newblock Monocular 3d reconstruction of interacting handsvia collision-aware factorized refinements.
\newblock In {\em International Conference on 3D Vision}, 2021.

\bibitem{ray_casting}
S.~D. Roth.
\newblock Ray casting for modeling solids.
\newblock {\em Computer Graphics and Image Processing}, 18(2):109--144, 1982.

\bibitem{Smith_Constraining}
B.~Smith, C.~Wu, H.~Wen, P.~Peluse, Y.~Sheikh, J.~K. Hodgins, and T.~Shiratori.
\newblock Constraining dense hand surface tracking with elasticity.
\newblock {\em ACM Trans. Graph.}, 39(6), nov 2020.

\bibitem{spurr2021self}
A.~Spurr, A.~Dahiya, X.~Wang, X.~Zhang, and O.~Hilliges.
\newblock Self-supervised 3d hand pose estimation from monocular rgb via contrastive learning.
\newblock In {\em Proceedings of the IEEE/CVF International Conference on Computer Vision}, pages 11230--11239, 2021.

\bibitem{spurr2018cvpr}
A.~Spurr, J.~Song, S.~Park, and O.~Hilliges.
\newblock Cross-modal deep variational hand pose estimation.
\newblock In {\em CVPR}, 2018.

\bibitem{Taylor_Articulated}
J.~Taylor, V.~Tankovich, D.~Tang, C.~Keskin, D.~Kim, P.~Davidson, A.~Kowdle, and S.~Izadi.
\newblock Articulated distance fields for ultra-fast tracking of hands interacting.
\newblock {\em ACM Trans. Graph.}, 36(6), nov 2017.

\bibitem{RGB2Hands}
J.~Wang, F.~Mueller, F.~Bernard, S.~Sorli, O.~Sotnychenko, N.~Qian, M.~A. Otaduy, D.~Casas, and C.~Theobalt.
\newblock Rgb2hands: Real-time tracking of 3d hand interactions from monocular rgb video.
\newblock {\em ACM Trans. Graph.}, 39(6), nov 2020.

\bibitem{Xiong_a2j}
F.~Xiong, B.~Zhang, Y.~Xiao, Z.~Cao, T.~Yu, J.~Zhou, and J.~Yuan.
\newblock A2j: Anchor-to-joint regression network for 3d articulated pose estimation from a single depth image.
\newblock In {\em 2019 IEEE/CVF International Conference on Computer Vision (ICCV)}, pages 793--802, Los Alamitos, CA, USA, nov 2019. IEEE Computer Society.

\bibitem{disentagling_hands_linlin}
L.~Yang and A.~Yao.
\newblock Disentangling latent hands for image synthesis and pose estimation.
\newblock pages 9869--9878, 06 2019.

\bibitem{yu2023acr}
Z.~Yu, S.~Huang, F.~Chen, T.~P. Breckon, and J.~Wang.
\newblock Acr: Attention collaboration-based regressor for arbitrary two-hand reconstruction.
\newblock In {\em Proceedings of the IEEE/CVF Conference on Computer Vision and Pattern Recognition (CVPR)}, June 2023.

\bibitem{Zhang2021twohand}
B.~Zhang, Y.~Wang, X.~Deng, Y.~Zhang, P.~Tan, C.~Ma, and H.~Wang.
\newblock Interacting two-hand 3d pose and shape reconstruction from single color image.
\newblock In {\em International Conference on Computer Vision (ICCV)}, 2021.

\bibitem{zhao_recover_3d}
R.~Zhao, Y.~Wang, and A.~Martinez.
\newblock A simple, fast and highly-accurate algorithm to recover 3d shape from 2d landmarks on a single image, 2016.

\bibitem{zb2017hand}
C.~Zimmermann and T.~Brox.
\newblock Learning to estimate 3d hand pose from single rgb images.
\newblock Technical report, arXiv:1705.01389, 2017.
\newblock https://arxiv.org/abs/1705.01389.

\end{thebibliography}
